\documentclass[conference]{IEEEtran}
\IEEEoverridecommandlockouts

\usepackage{cite}
\usepackage{amsmath,amssymb,amsfonts}
\usepackage{algorithmic}
\usepackage{graphicx}
\usepackage{textcomp}
\usepackage{xcolor}

\usepackage[T1]{fontenc}
\usepackage{multirow}
\usepackage{bm}
\usepackage{babel}
\usepackage{float}
\usepackage{booktabs}
\usepackage{url}
\usepackage{hyperref}

\usepackage{amsmath, amssymb, amsfonts}
\usepackage[format=plain,labelformat=simple,labelsep=period,font=small,compatibility=false,skip=3pt]{caption}
\usepackage[font=footnotesize,skip=3pt,subrefformat=parens]{subcaption}

\def\BibTeX{{\rm B\kern-.05em{\sc i\kern-.025em b}\kern-.08em
    T\kern-.1667em\lower.7ex\hbox{E}\kern-.125emX}}
    
\begin{document}

\title{Analyzing the Impact of Low-Rank Adaptation for Cross-Domain Few-Shot Object Detection in Aerial Images}

\author{
    \IEEEauthorblockN{Hicham TALAOUBRID}
    \IEEEauthorblockA{
        \textit{L2TI \& COSE} \\
        \textit{Université Sorbonne Paris Nord} \\
        hicham.talaoubrid1@edu.univ-paris13.fr
    }
    \and
    \IEEEauthorblockN{Anissa MOKRAOUI}
    \IEEEauthorblockA{
        \textit{L2TI} \\
        \textit{Université Sorbonne Paris Nord} \\
        anissa.mokraoui@univ-paris13.fr
    }
    \and
    \IEEEauthorblockN{Ismail BEN AYED}
    \IEEEauthorblockA{
        \textit{LIVIA, ETS} \\
        \textit{Montreal, Canada} \\
        Ismail.BenAyed@etsmtl.ca
    }
    \and
    \IEEEauthorblockN{Axel PROUVOST}
    \IEEEauthorblockA{
        \textit{IMT Mines Alès, France} \\
        axel.prouvost@etu.mines-ales.fr
    }
    \and
    \IEEEauthorblockN{Sonimith HANG}
    \IEEEauthorblockA{
        \textit{IMT Mines Alès, France} \\
        sonimith.hang@etu.mines-ales.fr
    }
    \and
    \IEEEauthorblockN{Monit KORN}
    \IEEEauthorblockA{
        \textit{IMT Mines Alès, France} \\
        monit.korn@etu.mines-ales.fr
    }
    \and
    \IEEEauthorblockN{Rémi HARVEY}
    \IEEEauthorblockA{
        \textit{COSE, Montmagny, France} \\
        remi.harvey@cose.fr
    }
}

\maketitle

\begin{abstract}
This paper investigates the application of Low-Rank Adaptation (LoRA) to small models for cross-domain few-shot object detection in aerial images. Originally designed for large-scale models, LoRA helps mitigate overfitting, making it a promising approach for resource-constrained settings. We integrate LoRA into DiffusionDet, and evaluate its performance on the DOTA and DIOR datasets. Our results show that LoRA applied after an initial fine-tuning slightly improves performance in low-shot settings (e.g., 1-shot and 5-shot), while full fine-tuning remains more effective in higher-shot configurations. These findings highlight LoRA’s potential for efficient adaptation in aerial object detection, encouraging further research into parameter-efficient fine-tuning strategies for few-shot learning. Our code is available here: \url{https://github.com/HichTala/LoRA-DiffusionDet}
\end{abstract}


\section{Introduction}

The last few years have seen remarkable improvement in large models. Particularly in natural language processing and computer vision \cite{touvron2023llama} \cite{radford2021learning}. Parameter-efficient fine-tuning methods have been developed to train these very large models for simpler tasks, without the need to train tens of billions of parameters. One of these is Low Rank Adaptation (LoRA), which, by injecting low rank matrices while freezing the model's pre-training weights, considerably reduces the number of parameters to be trained in the model. In this way, LoRA helps to limit the overfitting of large models by accelerating their convergence. But its potential in models 100 to 1000 times smaller in terms of parameters remains rather unexplored, particularly in a context of cross domain few-shot object detection, where the overfitting remains the main difficulty.

Few-shot Object Detection (FSOD) is a challenging task that aims to detect objects from novel categories using only a few labeled examples. When applied across domains, the problem becomes even more complex due to significant distribution shifts between source and target domains. Traditional fine-tuning approaches often struggle in such scenarios as they tend to overfit to the limited training data, especially in smaller models with fewer parameters. This challenge is even greater in cross-domain contexts, where the model needs to generalize to new domains with a minimum of supervision.

Aerial images present an additional complexity for object detection. These images often contain numerous small objects densely distributed over the scene, as well as significant variations in scale between classes, orientation, and illumination  \cite{lejeune2022improving}. These characteristics make aerial images a particularly demanding domain for cross-domain few-shot object-detection. The main problem remains overfitting in such a scenario. In this work, we investigate the application of LoRA to small models, using the COCO \cite{lin2014microsoft} dataset as the source domain and the DOTA \cite{xia2018dota} and DIOR \cite{li2020object} datasets as target domains. Both datasets are widely recognized references for aerial image analysis \cite{chen2024object} \cite{leng2024recent}.

The motivation behind this work is to address overfitting in cross-domain few-shot object detection, particularly for aerial images. We explore Low-Rank Adaptation (LoRA) \cite{edward2021lora}, a technique designed for large models, to improve generalization in smaller architectures like DiffusionDet \cite{chen2022diffusiondet}, which has about millions of parameters. DiffusionDet has shown effectiveness in detecting small objects, making it a good choice for aerial images. We compare DiffusionDet with and without LoRA, testing two strategies: (1) direct LoRA application and (2) LoRA after intermediate fine-tuning. Using the DOTA and DIOR datasets, we evaluated LoRA's ability to reduce overfitting and improve generalization, offering insight into its potential for efficient and robust aerial object detection.

\section{Related Work}
\subsection{Low-Rank Adaptation (LoRA) for Efficient Fine-Tuning}
LoRA \cite{edward2021lora} is a parameter-efficient fine-tuning method that adapts pre-trained models to new tasks by injecting low-rank decomposition matrices into existing weight layers. Instead of updating all parameters, LoRA introduces small trainable matrices into transformer layers, significantly reducing memory usage and computational costs while preserving the original model weights. Initially developed for natural language processing (NLP), LoRA has been extended to computer vision, enabling efficient adaptation of large-scale models to new domains with limited labeled data.

In the context of aerial images, LoRA has been applied to vision transformers trained on the DOTA \cite{xia2018dota} dataset, demonstrating its effectiveness in transfer learning for tasks such as object detection \cite{wang2023lora}. These studies highlight LoRA's ability to adapt large-scale models, such as ViT and Swin Transformer, to new aerial datasets without requiring full fine-tuning. Beyond aerial images, LoRA has also been successfully applied to Vision-Language Models (VLMs) like CLIP \cite{radford2021learning} and GLIP \cite{li2021grounded}, enabling task-specific adaptation without modifying the backbone network.

However, most of the research on LoRA has focused on large-scale models, where the overall trainable capacity remains high despite parameter reduction. Its application to smaller object detection models, which have significantly fewer parameters, remains largely unexplored. This is particularly relevant in cross-domain few-shot scenarios, where models must generalize from very limited samples and domain shifts are prevalent.

\subsection{Few-Shot and Cross-Domain Object Detection}
FSOD focuses on detecting novel object categories using only a small amount of labeled data. Two primary approaches dominate the literature \cite{kohler2021few}: meta-learning \cite{finn2017model} and fine-tuning-based methods. Meta-learning trains models to quickly adapt to new classes by learning transferable representations, while fine-tuning adapts pre-trained models to novel categories using limited labeled data. Although meta-learning excels at learning generalizable features, fine-tuning remains the dominant strategy, particularly in cross-domain scenarios where shifts in data distribution introduce additional complexity \cite{xiong2023cd}, \cite{lee2022rethinking}.

Cross-domain object detection extends FSOD by requiring models to generalize not only to new classes but also to different data distributions. The most common approach is fine-tuning, where a model pre-trained on a source domain is adapted to a target domain using a limited number of labeled examples.

Aerial object detection presents unique challenges due to variations in image resolution, sensor types, and environmental conditions across datasets. To address these domain shifts, researchers have explored techniques such as feature alignment \cite{tzeng2017adversarial} and multi-scale representation learning \cite{lin2017feature}, which aim to bridge the gap between satellite images and drone-captured images. However, these methods often require extensive adaptation, increasing computational complexity and limiting their practicality in resource-constrained scenarios.

\subsection{DiffusionDet}
DiffusionDet \cite{chen2022diffusiondet} is an object detection framework that formulates detection as a denoising diffusion process. Unlike traditional detectors that rely on handcrafted components like anchor boxes or region proposal networks, DiffusionDet directly predicts object bounding boxes and categories by iteratively refining noisy proposals. This approach has proven particularly effective for detecting small objects, making it well-suited for challenging domains such as aerial images, where objects are often densely distributed and vary significantly in scale.

Recent work has explored adapting DiffusionDet to few-shot object detection (FSOD) and cross-domain scenarios. For instance, \cite{lejeune2024improving} demonstrated that DiffusionDet can be fine-tuned for few-shot or cross-domain settings by leveraging its iterative refinement process to generalize better to categories with limited labeled data. However, these adaptations often require extensive fine-tuning, which can lead to overfitting, especially in resource-constrained settings with limited labeled data.

Our work builds on these advancements by integrating Low-Rank Adaptation (LoRA) into DiffusionDet, enabling efficient adaptation to cross-domain few-shot object detection. By injecting low-rank matrices into DiffusionDet’s architecture, we aim to reduce overfitting while maintaining the model’s ability to generalize across domains. This approach extends prior efforts by addressing overfitting challenges associated with fine-tuning DiffusionDet, particularly in the context of aerial images where domain shifts and data scarcity are prevalent.

\subsection{Other Detectors in Cross-Domain Few-Shot Scenarios}
Beyond DiffusionDet, several other detectors have been adapted for cross-domain few-shot object detection, each addressing the challenges of domain shifts and limited labeled data. For instance, CD-ViTO \cite{fu2024crossdomain} leverages vision transformers combined with cross-domain alignment techniques to improve generalization across diverse datasets. By integrating domain-adversarial training and feature alignment modules, CD-ViTO effectively reduces domain discrepancies while maintaining high detection accuracy in few-shot settings. Other approaches include Meta-Det \cite{yan2019meta}. These methods showcase the variety  of strategies available for tackling cross-domain few-shot detection. It would be interesting to extend our studies to include such models, further exploring the potential of LoRA in diverse architectures and settings.

\section{Challenge of Cross-Domain Few-Shot Object Detection in Aerial Images}
Aerial images present a unique set of challenges for object detection \cite{lejeune2022improving}, making them an especially demanding domain for few-shot learning and cross-domain adaptation. Unlike ground-level images, aerial scenes often contain numerous small objects, such as vehicles, buildings, or infrastructure components, which are densely distributed across the image. The small size and high density of these objects make them difficult to detect and localize accurately, even for state-of-the-art models. Additionally, aerial images exhibit significant variations in scale, orientation, and lighting conditions, further complicating the detection task. These challenges are exacerbated in few-shot settings, where models must learn to detect novel object categories with only a handful of examples, and cross-domain adaptation introduces even further complexity. In this context, the risk of overfitting is particularly high, as small models may struggle to capture the diverse and intricate patterns present in aerial scenes.

In this context, foundation models, including Vision-Language Models (VLMs) \cite{oquab2023dinov2} \cite{chen2024florencevl}, are less suitable for aerial object detection. Beyond the practical limitations of deploying such models in resource-constrained environments, their training data often focuses on horizontal perspectives, resulting in representations that are biased toward ground-level views. For example, in the embedding space of a VLM, the word "car" is more likely to be associated with horizontal representations of cars rather than their aerial counterparts. This misalignment between the model’s learned representations and the unique characteristics of aerial images limits their effectiveness in this domain.

Cross-domain few-shot object detection adds an additional level of complexity. In classic few-shot object detection, the model has access to a sufficient number annotated image of base classes from the same domain and aims to generalize to novel classes with limited examples. In contrast, cross-domain few-shot detection involves adapting to a target domain where all classes are novel, using a source domain with abundant data. This scenario requires the model to bridge significant domain shifts while learning from scarce labeled data.

\begin{figure}
    \centering
    \includegraphics[width=0.5\textwidth]{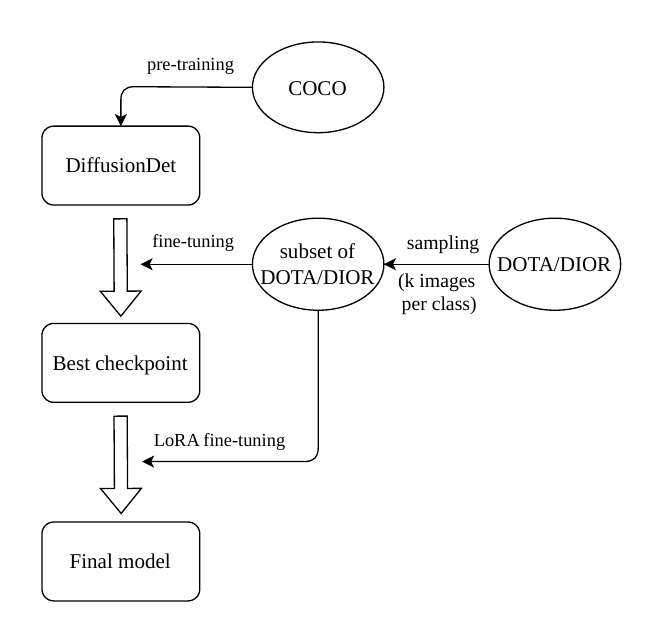}
    \caption{Training Pipeline for DiffusionDet with LoRA After Intermediate Fine-Tuning.}
    \label{fig:lora}
\end{figure}

\begin{table*}[t!]
\centering
\caption{Object etection results (mAP) of DiffusionDet model pretrained on COCO \cite{lin2014microsoft} and fine-tuned on DIOR \cite{li2020object} and DOTA \cite{xia2018dota} datasets in cross-domain few-shot settings. We compare the baseline (no LoRA), LoRA with different ranks (4, 8, 32, 128), and LoRA applied after intermediate fine-tuning. Best results per shot configuration are bold.}
\label{tab:results}
\resizebox{.8\textwidth}{!}{
\begin{tabular}{@{}cc|c|cccc|cccc@{}}
\toprule
\multirow{2}{*}{Dataset} & \multirow{2}{*}{Shots} & \multirow{2}{*}{\begin{tabular}[c]{@{}c@{}}Baseline\\ (no LoRA)\end{tabular}} & \multicolumn{4}{c|}{LoRA} & \multicolumn{4}{c}{LoRA after a Fine-Tuning} \\
\cmidrule(lr){4-7} \cmidrule(lr){8-11}
 & & & 4 & 8 & 32 & 128 & 4 & 8 & 32 & 128 \\
\midrule
\multirow{4}{*}{DIOR} & 1 & 10.66 & 7.32 & 6.83 & 7.51 & 6.52 & 11.48 & 11.58 & \textbf{11.64} & 11.57 \\
 & 5 & 31.29 & 24.14 & 24.84 & 24.02 & 24.45 & 32.40 & 32.2 & 32.35 & \textbf{32.45} \\
 & 10 & \textbf{41.50} & 34.72 & 34.25 & 33.91 & 33.23 & 40.64 & 40.68 & 40.81 & 41.18 \\
 & 50 & \textbf{59.71} & 56.43 & 53.41 & 53.24 & 56.47 & 57.72 & 57.74 & 57.78 & 57.70 \\
\midrule
\multirow{4}{*}{DOTA} & 1 & 4.23 & 1.86 & 1.81 & 1.81 & 1.70 & 4.89 & 4.85 & 4.84 & \textbf{4.97} \\
 & 5 & 22.52 & 15.17 & 14.83 & 15.15 & 14.73 & 22.75 & 22.83 & \textbf{22.91} & 22.85 \\
 & 10 & \textbf{32.77} & 25.12 & 24.54 & 25.07 & 25.06 & 32.23 & 32.33 & 32.30 & 32.14 \\
 & 50 & \textbf{49.17} & 42.90 & 42.07 & 42.50 & 42.15 & 47.90 & 47.99 & 48.03 & 47.94 \\
\bottomrule
\end{tabular}}
\end{table*}
\section{Methodology}
\label{sec: metho}
Our methodology evaluates the effectiveness of LoRA \cite{edward2021lora} for cross-domain few-shot object detection using the DiffusionDet \cite{chen2022diffusiondet} framework. We begin with a pre-trained
DiffusionDet model on the COCO \cite{lin2014microsoft} dataset, which serves as the source domain, and fine-tune it on reduced subsets of the DOTA \cite{xia2018dota} and DIOR \cite{li2020object} datasets as target domains. These subsets are carefully curated to simulate a few-shot regime, ensuring that the model is trained with limited labeled data. The backbone used for DiffusionDet is ResNet50 \cite{he2016deep}, providing a balance between computational efficiency and feature extraction capability.

First, to simulate a few-shot setting, we randomly sampled multiple subsets from the DOTA and DIOR datasets for each shot configuration. Averaging the results across these subsets accounts for selection variability, reduces dataset bias, and ensures a fair evaluation of the model’s generalization ability.

To establish a baseline, we first fine-tune the pre-trained DiffusionDet model on the reduced subsets of DOTA and DIOR without freezing any parameters. The results from this baseline experiment serve as a reference point for evaluating the effectiveness of LoRA in the subsequent steps.

To evaluate the effectiveness of LoRA in cross-domain few-shot object detection, we designed a series of experiments comparing two key approaches, as described below.\\
1) In the first one, we explore the direct application of LoRA to the pre-trained DiffusionDet model. By injecting low-rank matrices into the model’s architecture and freezing the original weights, we significantly reduce the number of trainable parameters. This approach aims to mitigate overfitting while maintaining the model’s ability to generalize across domains. The model is then fine-tuned on the reduced subsets of DOTA and DIOR, and its performance is evaluated on a separate validation set. \\
2) In this second approach, illustrated in Fig. \ref{fig:lora}, we first fine-tune the pre-trained DiffusionDet model on the reduced subsets of DOTA and DIOR until it reaches a checkpoint with optimal performance on the validation set, just before overfitting occurs. We then apply LoRA to this checkpoint, freezing the original weights and fine-tuning only the low-rank matrices. The idea behind this two-stage approach is to first push the model to its limits, approaching the point of overfitting, and then use LoRA to continue training without the risk of overfitting. By applying LoRA to this checkpoint, we freeze the original weights and fine-tune only the low-rank matrices, allowing the model to adapt further while maintaining generalization. This strategy leverages the benefits of both full fine-tuning and parameter-efficient adaptation, achieving a better balance between performance and robustness.

\section{Experiments and Results}
To evaluate the effectiveness of LoRA in cross-domain few-shot object detection, we conducted a series of experiments using the DiffusionDet framework. Our experiments were designed to assess the impact of LoRA on model performance, overfitting, and generalization across domains, particularly in the challenging context of aerial images.

We used DiffusionDet model pre-trained on the COCO dataset, using the weights provided by the original authors \cite{chen2022diffusiondet}. For the target domains, we selected the DOTA and DIOR datasets, which were converted to COCO format\footnote{They can be found here: \url{https://huggingface.co/datasets/HichTala/dota},\\ \url{https://huggingface.co/datasets/HichTala/dior}.}. To simulate a few-shot setting, we randomly selected $k$ images per class for training, where $k$ represents the number of shots. To ensure a fair comparison and account for variability in the selection of images, we repeated each experiment 5 times and reported the average results. Given that DOTA images often contain more than 100 objects, we set the maximum detection threshold to 300 in the pycocoapi evaluation toolkit \cite{cocoapi}.

All experiments were conducted over 300 epochs, following the training protocol established by the original DiffusionDet authors. We evaluated the model performance using the mean average precision (mAP) at an IoU threshold of 0.5, which is a standard metric for object detection tasks. For the baseline, we fine-tuned the pre-trained DiffusionDet model on the few-shot subsets of DOTA and DIOR without freezing any parameters. As described in section \ref{sec: metho}, for the LoRA-based experiments, we explored two approaches: applying LoRA directly to the pre-trained model and applying LoRA to the best checkpoint obtained from the baseline fine-tuning. In the latter approach, we selected the checkpoint with the highest validation performance after 300 epochs and fine-tuned it further using LoRA.

To investigate the impact of rank selection on LoRA’s performance, we tested four different ranks: 4, 8, 32, and 128. The results of these experiments are presented in Tab. \ref{tab:results}, which compares the performance of the baseline, direct LoRA application, and LoRA after intermediate fine-tuning across different ranks. By averaging results across multiple runs, we ensure a robust evaluation of LoRA’s effectiveness in mitigating overfitting and improving generalization in cross-domain few-shot object detection.

Across both datasets, the baseline outperforms direct LoRA application in all shot configurations. However, LoRA applied after intermediate fine-tuning shows improvements, particularly in low-shot settings. On DIOR, the best mAP of 11.64 (rank 32) is achieved in the 1-shot setting, while on DOTA, the best mAP of 4.97 (rank 128) is achieved. Similarly, in the 5-shot setting, the best mAPs are 32.45 (rank 128) on DIOR and 22.91 (rank 32) on DOTA. In higher-shot settings, the baseline remains competitive, but LoRA after fine-tuning closely matches its performance.

LoRA after intermediate fine-tuning slightly improves performance in low-shot settings, while the baseline remains strong in higher-shot configurations. The choice of rank in LoRA has a moderate impact on performance, with lower ranks (e.g., 4, 8) often performing comparably to higher ranks (e.g., 32, 128).

\section{Discussion}
The experimental results demonstrate that LoRA, particularly when applied after intermediate fine-tuning, is a promising approach for cross-domain few-shot object detection. This improvement, 	although minimal, suggests that efficient parameter fine-tuning could be a viable alternative to full fine-tuning, particularly in resource-constrained environments.

The choice of rank in LoRA has a moderate impact on performance, with lower ranks (e.g., 4, 8) often performing comparably to higher ranks (e.g., 32, 128). This indicates that lower ranks may suffice for many applications. However, the baseline’s strong performance in higher-shot configurations underscores the importance of full fine-tuning when sufficient data is available. These results underline the need for a balanced approach, adjusting the fine-tuning strategy to the specific requirements of the task and dataset, and could be the subject of further study.

While our approach shows promise, it is not without limitations. The performance of LoRA depends on the quality of the initial fine-tuning. Additionally, our experiments are limited to DiffusionDet and two aerial datasets; extending this approach to other architectures and domains could yield further insights. Future work could explore combining LoRA with other few-shot learning techniques to enhance its effectiveness.

It is worth noting that the cross-domain scenario explored in this work—adapting from natural images (COCO) to aerial images (DOTA and DIOR), remains particularly challenging due to the significant differences in perspective, scale, and object appearance. A simpler yet equally interesting scenario would involve adapting between aerial images, such as from DOTA to DIOR or vice versa, where the domain shifts are less extreme. One could also explore adapting aerial images across different environments, seasons, or lighting conditions. Such experiments could provide valuable insights into the robustness and versatility of LoRA in less extreme domain shifts.

\section{Conclusion}
In this work, we investigated the application of LoRA to DiffusionDet model for cross-domain few-shot object detection, with a focus on the challenging domain of aerial images. Using the DiffusionDet framework, we evaluated the effectiveness of LoRA in mitigating overfitting and improving generalization across the DOTA and DIOR datasets. Our experiments compared three approaches: (1) baseline fine-tuning, (2) direct LoRA application, and (3) LoRA applied after intermediate fine-tuning. The results demonstrated that while the baseline outperformed direct LoRA application, LoRA after intermediate fine-tuning achieved competitive performance, particularly in low-shot settings (e.g., 1-shot and 5-shot). This highlights LoRA’s potential to balance adaptation and generalization, especially when combined with an initial fine-tuning phase.

\bibliographystyle{IEEEtran}
\bibliography{references}

\end{document}